\documentclass[letterpaper, 10 pt, conference]{ieeeconf}  

\usepackage{graphicx}

\IEEEoverridecommandlockouts

\usepackage{cite}

\title{\LARGE \bf
AsyncShield: A Plug-and-Play Edge Adapter for Asynchronous Cloud-based VLA Navigation
}

\author{Kai Yang$^{1}$, Zedong Chu$^{1\dagger}$, Yingnan Guo$^{1}$, Zhengbo Wang$^{1}$, Shichao Xie$^{1}$, \\
Yanfen Shen$^{1}$, Xiaolong Wu$^{1}$, Xing L\"{u}$^{2}$, and Mu Xu$^{1}$
\thanks{$^{\dagger}$Corresponding author.}%
\thanks{$^{1}$Amap, Alibaba Group, Beijing, China. Emails: \{yk496472, chuzedong.czd\}@alibaba-inc.com}%
\thanks{$^{2}$Beijing Jiaotong University, Beijing, China.}%
}

\usepackage{amsmath, amssymb}

\usepackage{booktabs} 
\usepackage{multirow} 
\usepackage[font=small, labelsep=colon, justification=justified, singlelinecheck=false]{caption} 

\begin{document}

\maketitle

\begin{abstract}
While Vision-Language-Action (VLA) models have been demonstrated possessing strong zero-shot generalization for robot control, their massive parameter sizes typically necessitate cloud-based deployment. However, cloud deployment introduces network jitter and inference latency, which can induce severe spatiotemporal misalignment in mobile navigation under continuous displacement, so that the stale intents expressed in past ego frames may become spatially incorrect in the current frame and lead to collisions. To address this issue, we propose AsyncShield, a plug-and-play asynchronous control framework. AsyncShield discards traditional black-box time-series prediction in favor of a deterministic physical white-box spatial mapping. By maintaining a temporal pose buffer and utilizing kinematic transformations, the system accurately converts temporal lag into spatial pose offsets to restore the VLA's original geometric intent. To balance intent restoration fidelity and physical safety, the edge adaptation is formulated as a constrained Markov decision process (CMDP). Solved via the PPO-Lagrangian algorithm, a reinforcement learning adapter dynamically trades off between tracking the VLA intent and responding to high-frequency LiDAR obstacle avoidance hard constraints. Furthermore, benefiting from a standardized universal sub-goal interface, domain randomization, and perception-level adaptation via Collision Radius Inflation, AsyncShield operates as a lightweight, plug-and-play module. Simulation and real-world experiments demonstrate that, without fine-tuning any cloud-based foundation models, the framework exhibits zero-shot and robust generalization capabilities, effectively improving the success rate and physical safety of asynchronous navigation.
\end{abstract}


\section{INTRODUCTION}
\label{sec:intro}
While Vision-Language-Action (VLA) models have been demonstrated possessing impressive zero-shot generalization capabilities in robotic manipulation~\cite{zhang2024uni,brohan2023rt2visionlanguageactionmodelstransfer,octomodelteam2024octoopensourcegeneralistrobot,kim2024openvlaopensourcevisionlanguageactionmodel} and robotic navigation~\cite{chu2026abotn0technicalreportvla,hu2025astranavworldworldmodelforesight,xiang2025navr2dualrelationreasoninggeneralizable,liu2025navforeseeunifiedvisionlanguageworld,chen2025socialnavtraininghumaninspiredfoundation,xue2026omninavunifiedframeworkprospective}, their massive parameter sizes usually necessitate cloud-based deployment. This inevitably introduces systemic cloud-to-edge latency. In dynamic environments, inference suffers from significant latency while control must operate in real-time. Consequently, semantic information frequently corresponds to past states but is utilized as current, thereby triggering a systemic temporal misalignment between ``thinking'' and ``control''~\cite{huang2026ticvlathinkincontrolvisionlanguageactionmodel}. How to safely transfer the powerful capabilities of cloud-based large models to highly dynamic mobile navigation tasks remains an urgent challenge.

Existing asynchronous control frameworks (e.g., RTC~\cite{black2025realtimeexecutionactionchunking} and A2C2~\cite{sendai2025leaveobservationbehindrealtime}) predominantly focus on temporal alignment, mitigating latency via smooth action chunk splicing or local residual correction. While effective in fixed-base manipulator operations, these methods face a fundamental logic failure when transferred to mobile robots undergoing continuous, large-scale displacements: blindly and smoothly fitting an outdated path prevents the robot from adequately responding to dynamic obstacles. Furthermore, end-to-end delay-injection training methods struggle to cope with long-tail, irregular network jitter in the real world; meanwhile, traditional control frameworks based on black-box time-series prediction also prove extremely fragile under extreme communication latency, easily leading to catastrophic task failures.

To address these issues, we propose the AsyncShield, a framework that transforms traditional asynchronous control based on black-box time prediction into a deterministic physical white-box mapping and a safe execution closed loop. We maintain a historical temporal pose buffer at the edge. Upon receiving outdated VLA commands, the system employs an analytical $SE(2)$ kinematic transformation to eliminate the cloud-to-edge latency misalignment, thereby restoring the VLA's true geometric intent. To balance the restoration fidelity of the VLA intent and physical safety, we formulate the edge adaptation as a CMDP. Through the PPO-Lagrangian algorithm~\cite{achiam2017constrained}, the reinforcement learning adapter can adaptively trade off between tracking the true VLA intent and responding to the high-frequency LiDAR obstacle avoidance hard constraints.

Furthermore, AsyncShield is designed as an extremely lightweight, independent edge adaptation architecture, emphasizing ``plug-and-play'' capability. Unlike existing solutions that require deep modifications or fine-tuning of the cloud-based foundation models, we standardize the cloud VLA output via interpolation resampling into $5$ local waypoints with a $20$~cm spacing as the input to the adapter. Based on this, the policy network solely outputs a Universal Local Sub-goal. This standardized interface design, combined with various randomization training strategies, endows the system with plug-and-play generalization capabilities, seamlessly accommodating various cloud VLA models and generalizing across multiple heterogeneous robot chassis without any fine-tuning.

In summary, the main contributions of this paper are as follows:
    

\begin{itemize}
    \item \textbf{Cloud-based VLA Navigation Edge Adapter:} We propose the AsyncShield framework, which replaces black-box time prediction with deterministic white-box spatial mapping. Concurrently, by formulating a CMDP-based safe execution closed loop, it achieves an adaptive dynamic trade-off between the restoration of the large model's geometric intent and low-level, high-frequency obstacle avoidance.
    
    \item \textbf{Strong Plug-and-Play Capability:} We propose a lightweight, independent edge adaptation strategy with a standardized interface. Without requiring any fine-tuning of the cloud-based foundation models, AsyncShield can zero-shot seamlessly accommodate various cloud VLA models and reliably generalize across multiple heterogeneous mobile robot chassis.
\end{itemize}

\section{RELATED WORKS}

\subsection{Vision-Language-Action Models and Hierarchical Adaptation}
In recent years, VLA models (e.g., RT-2~\cite{brohan2023rt2visionlanguageactionmodelstransfer}, OmniVLA~\cite{hirose2025omnivlaomnimodalvisionlanguageactionmodel}, Green-VLA~\cite{apanasevich2026greenvlastagedvisionlanguageactionmodel}) have demonstrated formidable zero-shot generalization capabilities, gradually expanding into highly dynamic scenarios such as unmanned aerial vehicles (AutoFly~\cite{sun2026autoflyvisionlanguageactionmodeluav}), autonomous driving (Impromptu VLA~\cite{chi2025impromptu}), and dynamic object interaction (DynamicVLA~\cite{xie2026dynamicvla}). However, their massive parameter sizes lead to extremely high inference latency and network jitter. To alleviate the deployment bottleneck at the edge, existing research has primarily evolved along multiple trajectories. One approach lowers the computational threshold through model lightweighting, context compression, or post-training reinforcement learning fine-tuning (e.g., SmolVLA~\cite{shukor2025smolvla}, ContextVLA~\cite{jang2025contextvla}, SimpleVLA-RL~\cite{li2025simplevla}). Another constructs a ``fast-slow dual-system'' hierarchical architecture (e.g., Mobility VLA~\cite{chiang2024mobilityvlamultimodalinstruction}, IROS Dual-Process~\cite{lee2026irosdualprocessarchitecturerealtime}), or introduces world models with latent state spaces to enhance generalization (e.g., X-MOBILITY~\cite{liu2025x}). Particularly in the domain of cross-embodiment navigation, works such as X-Nav~\cite{wang2025x} explore the end-to-end distillation of massive expert policies, while CE-Nav~\cite{yang2025nav} goes a step further by successfully decoupling high-level geometric reasoning from low-level dynamic execution via a two-stage architecture. Similarly, ABot-Explorer~\cite{chen2026explorelikehumansautonomous} enhances high-level guidance through online scene graph memory construction. Nevertheless, the aforementioned works often implicitly rely on synchronous control assumptions. When faced with the irregular communication delays inherent in real-world cloud deployment, the closed-loop execution between high-level, low-frequency semantics and low-level, high-frequency physical execution is highly susceptible to fracture.  As a fully plug-and-play module, our AsyncShield circumvents communication and computational bottlenecks without necessitating any intervention in the internal weights of the cloud-based foundation model.

\subsection{Asynchronous Control and Latency-Aware Local Navigation}
Action Chunking techniques (e.g., ACT~\cite{zhao2023learningfinegrainedbimanualmanipulation}, Diffusion Policy~\cite{chi2024diffusionpolicyvisuomotorpolicy}) have been widely validated in fixed-base manipulator operations, and recent works like Mixture of Horizons~\cite{jing2025mixture} further explore adaptive fusion strategies for multi-horizon chunks. To address the asynchronous latency of mobile-based large models, existing strategies predominantly focus on ``temporal fitting and prediction.'' For instance, VLASH~\cite{tang2025vlashrealtimevlasfuturestateaware} relies on forward-state prediction to estimate the robot's future pose; DuoCore-FS~\cite{zou2025asynchronous} attempts to construct a latent representation buffer to bridge the dual-track system, whereas AsyncVLA~\cite{hirose2026asyncvla} concentrates on edge-side asynchronous adaptation manipulation via cloud-edge collaboration. Simultaneously, within low-level local perception modules, traditional frameworks (e.g., NavFormer~\cite{wang2024navformer}, V-STRONG~\cite{jung2024v}) and various reinforcement learning schemes (e.g., Hierarchical RL Nav~\cite{gao2025hierarchical}, Enhanced PPO~\cite{taheri2024deep}, Decentralized RL~\cite{lin2024decentralized}, APD for SRL~\cite{chen2024adaptive}) exhibit excellent environmental adaptability. However, when receiving expired commands from cloud-based large models, blind action chunk splicing along the temporal axis still leads to catastrophic spatial misalignment for mobile bases. This paper breaks the traditional paradigm of temporal sequence fitting by proposing ``Latency is Geometry.'' Our system directly utilizes analytical kinematic transformations within the $SE(2)$ space to precisely map temporal lag into spatial relative offsets within the ego-centric coordinate frame. Combined with off-the-shelf local reinforcement learning obstacle avoidance algorithms acting as low-level cost constraints, this physical white-box mapping eliminates the uncertainty brought by black-box predictions, achieving high-frequency, robust control free from manual parameter tuning.

\begin{figure*}[t]
    \centering
    \includegraphics[width=0.9\textwidth]{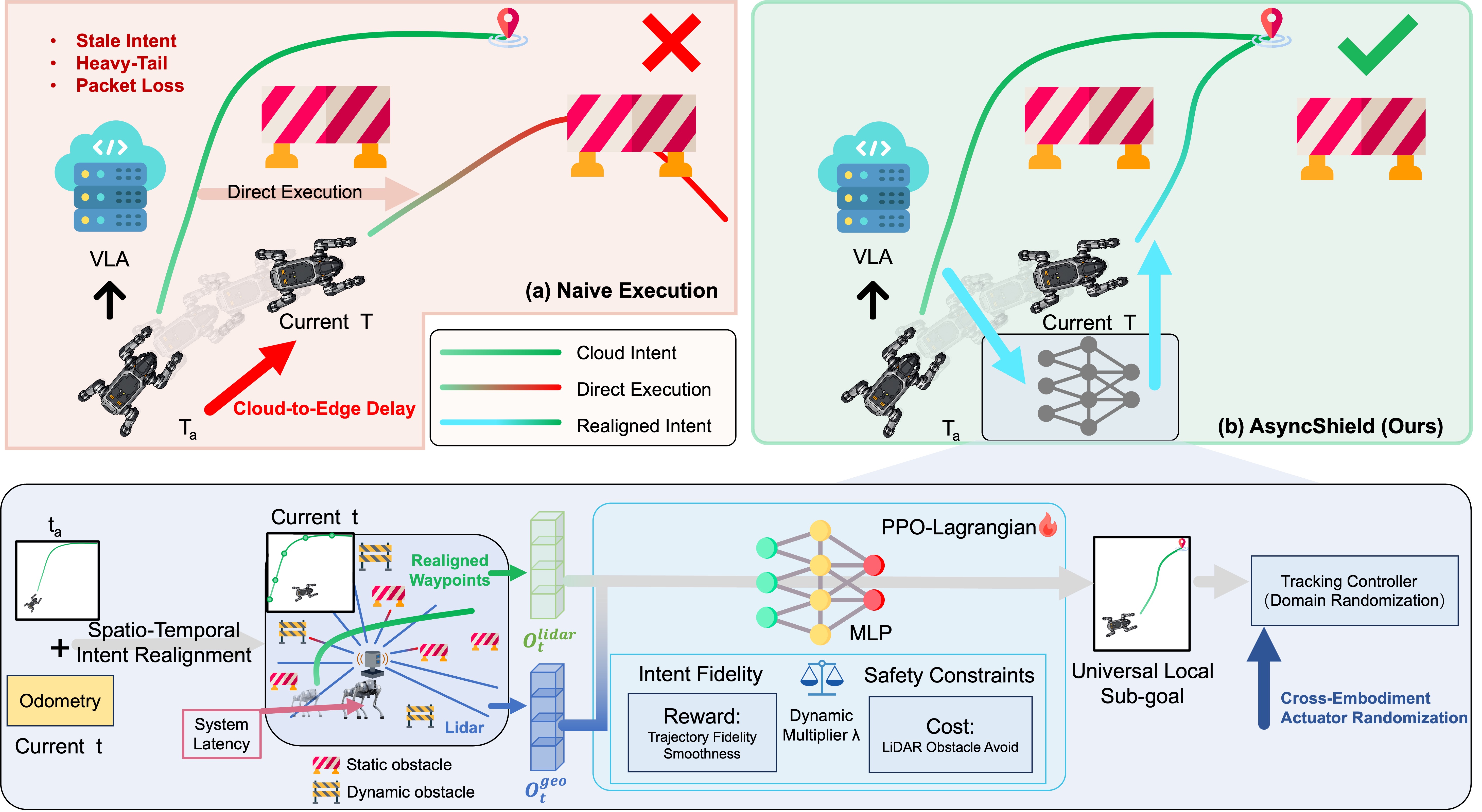}      
    \caption{\textbf{Overview of the AsyncShield framework.} \textbf{Top:} Behavioral comparison under network degradation. \textit{Naive Execution} blindly follows stale intents, leading to collisions, whereas AsyncShield safely bypasses obstacles. \textbf{Bottom:} The edge adaptation pipeline. The system first utilizes a temporal pose buffer to perform spatio-temporal intent realignment on delayed cloud waypoints. Subsequently, a policy optimized via \textit{PPO-Lagrangian} adaptively trades off between intent fidelity and hard safety constraints to output universal local sub-goals. Finally, by incorporating actuator domain randomization during training, the framework achieves plug-and-play generalization across embodiments.}
\end{figure*}
\section{METHODOLOGY}

The AsyncShield framework is designed to bridge the spatio-temporal gap between low-frequency, high-latency cloud VLA models and high-frequency, safety-critical edge execution. We formulate this asynchronous control problem as a constrained Markov decision process, where the edge adapter rectifies stale semantic intents through explicit geometric realignment and ensures physical safety via constrained policy optimization.

\subsection{Spatio-Temporal Intent Realignment}
To address displacement-induced misalignment during continuous motion, we discard implicit temporal fitting and instead employ an analytical $SE(2)$ transformation to map ``instruction lag'' into a deterministic ``spatial offset.''

\textbf{1) Temporal Pose Buffer:} The edge device maintains a circular buffer $\mathcal{B} = \{(t_k, \mathbf{T}_{W}^{O}(t_k))\}$, recording the robot's odometry (from world $W$ to ego frame $O$) at $f_{edge}$ Hz. When a VLA packet arrives with an anchor timestamp $t_a$, the system retrieves the corresponding historical pose $\mathbf{T}_{W}^{O}(t_a)$ via linear interpolation for translation and shortest-path angular interpolation for the heading in the $SE(2)$ space.

\textbf{2) Geometric Re-projection:} Let $\mathcal{P}^{A}=\{\bar{p}_{i}^{A}\}_{i=1}^{N}$ be the set of $N$ local waypoints generated by the VLA model in the Anchor Ego Frame at $t_a$. The realigned waypoints $\bar{p}_{i}^{E}(t)$ in the Current Ego Frame at time $t>t_{a}$ are computed analytically:
\begin{equation}
\overline{p}_{i}^{E}(t)=(T_{W}^{O}(t))^{-1}T_{W}^{O}(t_{a})\overline{p}_{i}^{A}
\end{equation}
where $\overline{p}$ denotes homogeneous coordinates. 


The spatial re-projection formula exclusively computes relative pose changes within the minimal delay window $\Delta t$. By establishing a new temporal anchor $t_a$ for each incoming VLA packet, this mechanism strictly confines odometry drift (e.g., from wheel slip) to a single communication cycle. Consequently, previous spatial alignment errors are instantly reset to zero upon receiving new waypoints, effectively preventing global divergence over time.


We define the edge adaptation as a CMDP tuple $(\mathcal{S, A, P, R, C}, \gamma, d)$ with transition dynamics $\mathcal{P}$ and discount factor $\gamma$, aiming to maximize the intent-tracking reward $J_R$ while keeping the expected safety cost $J_C$ below a threshold $d$.

\textbf{1) State and Action Space:} The state $s_t = [\mathbf{o}^{geo}_t, \mathbf{o}^{lidar}_t]$ integrates geometric and reactive features. $\mathbf{o}^{geo}_t \in \mathbb{R}^{10}$ consists of 5 look-ahead points sampled from the realigned path at $0.2\,\text{m}$ intervals. $\mathbf{o}^{lidar}_t \in \mathbb{R}^{144}$ provides 2D LiDAR proximity data. 
To ensure cross-embodiment compatibility, the action $a_t = (\Delta x_t, \Delta y_t)$ is defined as a Universal Local Sub-goal in the current ego frame, which is subsequently converted into velocity commands by a low-level controller.

\begin{table*}[t]
\centering
\small
\caption{Quantitative comparison under two network conditions. SR, CTE, and RER are computed over all 600 episodes (CTE is episode-weighted). RER is the percentage of time steps with $d_{\min} < d_{\text{risk}}$. TTG is computed on the shared-success set (intersection of success episodes across all compared methods; 120/600 for Ideal and 100/600 for Delayed).}
\label{tab:main_results}
\begin{tabular}{l cccc cccc}
\toprule
\multirow{2}{*}{\textbf{Method}} & \multicolumn{4}{c}{\textbf{Ideal (Fast Update)}} & \multicolumn{4}{c}{\textbf{Non-ideal (Mixed Degradation)}} \\
\cmidrule(lr){2-5} \cmidrule(lr){6-9}
& SR (\%) $\uparrow$ & CTE (m) $\downarrow$ & RER (\%) $\downarrow$ & TTG (s) $\downarrow$ & SR (\%) $\uparrow$ & CTE (m) $\downarrow$ & RER (\%) $\downarrow$ & TTG (s) $\downarrow$ \\
\midrule
\textbf{Ours} & \textbf{80.0} & 0.717 & \textbf{1.2} & 8.87 & \textbf{76.7} & \textbf{0.725} & \textbf{1.3} & \textbf{9.29} \\
A2C2                        & 56.7 & 0.937 & 1.5 & 8.41 & 43.3 & 1.146 & 1.7 & 9.64 \\
RTC                         & 40.0 & 0.673 & 3.0 & 8.18 & 30.0 & 1.178 & 4.0 & 9.81 \\
Naive                       & 20.0 & 1.175 & 5.2 & 7.92 & 16.7 & 1.272 & 5.5 & 10.25 \\
\bottomrule
\end{tabular}
\end{table*}

\textbf{2) Reward Design (Intent Fidelity):} The reward function $r_t$ focuses on trajectory fidelity and smoothness, independent of obstacle avoidance,
\begin{equation}
r_t = w_{flow}(\mathbf{a}_t^\top \hat{\tau}_t) - w_{cte}\text{tanh}(d^{cte}_t) - w_{smooth}\|\mathbf{a}_t - \mathbf{a}_{t-1}\|_2
\end{equation}
where $w_{flow}, w_{cte}, w_{smooth}$ are positive weight coefficients, and $\hat{\tau}_t$ is the local path unit tangent.To ensure gradient continuity, the cross-track error $d^{cte}_t$ is calculated via the point-to-line segment distance:
\begin{equation}
d^{cte}_t = \| \mathbf{p}_0 + t^*(\mathbf{p}_1 - \mathbf{p}_0) \|_2
\end{equation}


\begin{equation}
\quad t^* = \text{clip}\left(\frac{-\mathbf{p}_0^\top(\mathbf{p}_1-\mathbf{p}_0)}{\|\mathbf{p}_1-\mathbf{p}_0\|^2}, 0, 1\right)
\end{equation}
where $\mathbf{p}_0, \mathbf{p}_1$ are the immediate waypoints in the look-ahead window.

\textbf{3) Safety Constraints:} Physical safety is enforced as an independent cost $c_t$ based on the minimum LiDAR distance $d_{min}$. Given a safety radius $R_{safe}$, the cost is triggered as:
\begin{equation}
c_t = \mathbb{I}(d_{min} < R_{safe}) + \alpha \max(0, R_{safe} - d_{min})
\end{equation}
where $\mathbb{I}(\cdot)$ is the indicator function and $\alpha$ is a penalty scaling factor. 
We employ the \textbf{PPO-Lagrangian} algorithm~\cite{achiam2017constrained} to update a dual variable $\lambda$ that balances $J_R$ and $J_C$. When stale VLA intents pose collision risks, the surge in $\lambda$ forces the policy to prioritize safety advantages, resulting in proactive collision avoidance .

\subsection{Training Environment and Kinematic Domain Randomization}
To ensure zero-shot transferability, we train the AsyncShield in a highly stochastic simulated environment using the OmniSafe framework with extensive perturbations.

\textbf{1) Stochastic Environment Configuration:} Training is conducted in a $10\,\text{m} \times 10\,\text{m}$ workspace featuring both static and dynamic collision risks. Each episode is initialized with:
\begin{itemize}
    \item \textbf{Geometric Diversity:} 6 static and 6 dynamic obstacles with heterogeneous geometries (cylinders, polygons, and irregular shapes). The equivalent radius $R_{obs}$ is sampled from $\mathcal{U}(0.2, 2.0)\,\text{m}$.
    \item \textbf{Dynamic Interference:} Dynamic obstacles follow a random-walk model with linear velocities $v_{obs} \in [0.2, 1.0]\,\text{(m/s)}$ and angular velocities $\omega_{obs} \in [0.1, 1.0]\,\text{(rad/s)}$.
    \item \textbf{Asynchronous Disturbances:} We simulate irregular communication by sampling latencies $\delta \sim \mathcal{U}(0.3, 1.5)\,\text{s}$ and packet loss probabilities $p_{loss} \in [0, 0.2]$ , and intermittent transient outages (blackouts) lasting up to several seconds .
\end{itemize}


\textbf{2) Cross-Embodiment Actuator Randomization:} We decouple the policy from specific robot dynamics by randomizing the actuator response model. The transition from a sub-goal $a_t$ to the actual executed velocity $v_{act}$ is governed by acceleration constraints followed by a first-order lag system:
\begin{equation}
\begin{split}
    v_{act}(t) = \text{clip} \big( \tau v_{cmd}(t) + (1-\tau)v_{act}(t-1) \\
    + \eta_v, -v_{max}, v_{max} \big)
\end{split}
\end{equation}
where $v_{cmd}$ is the intended command derived from $a_t$, and $v_{max}$ is the velocity limit. The randomization parameters are sampled per episode to cover diverse chassis profiles:

\begin{itemize}
    \item \textbf{System Latency:} The first-order lag coefficient $\tau \sim \mathcal{U}(0.2, 0.9)$, representing varying motor response times.
    \item \textbf{Acceleration Constraints:} Maximum acceleration $a_{max} \sim \mathcal{U}(0.5, 1.5)\,\text{m/s}^2$ to simulate different power-to-weight ratios.
    \item \textbf{Stochastic Noise \& Bias:} Velocity-dependent Gaussian noise $\eta_v$ with intensity $\sigma \sim \mathcal{U}(0.05, 0.20)$ and systematic angular bias $b_{\omega} \sim \mathcal{U}(-0.05, 0.05)\,\text{rad/s}$.
\end{itemize}

This extensive kinematic perturbation ensures that the AsyncShield learns a dynamics-agnostic control law capable of seamless deployment on heterogeneous mobile platforms.

\section{EXPERIMENTS}

This section evaluates the performance of the AsyncShield framework under asynchronous cloud-edge control. We design extensive simulation experiments to answer the following questions: (1) Can AsyncShield maintain system robustness and navigational safety under varying network latencies? (2) What is the inherent trade-off between ``action smoothness'' and ``physical safety'' in mobile robot navigation? (3) How do the core modules---temporal pose buffer, RL adapter, and constrained optimization---contribute to the overall performance?

\subsection{Experimental Setup and Metrics}

\begin{figure*}[t]
    \centering
    \includegraphics[width=0.75\textwidth]{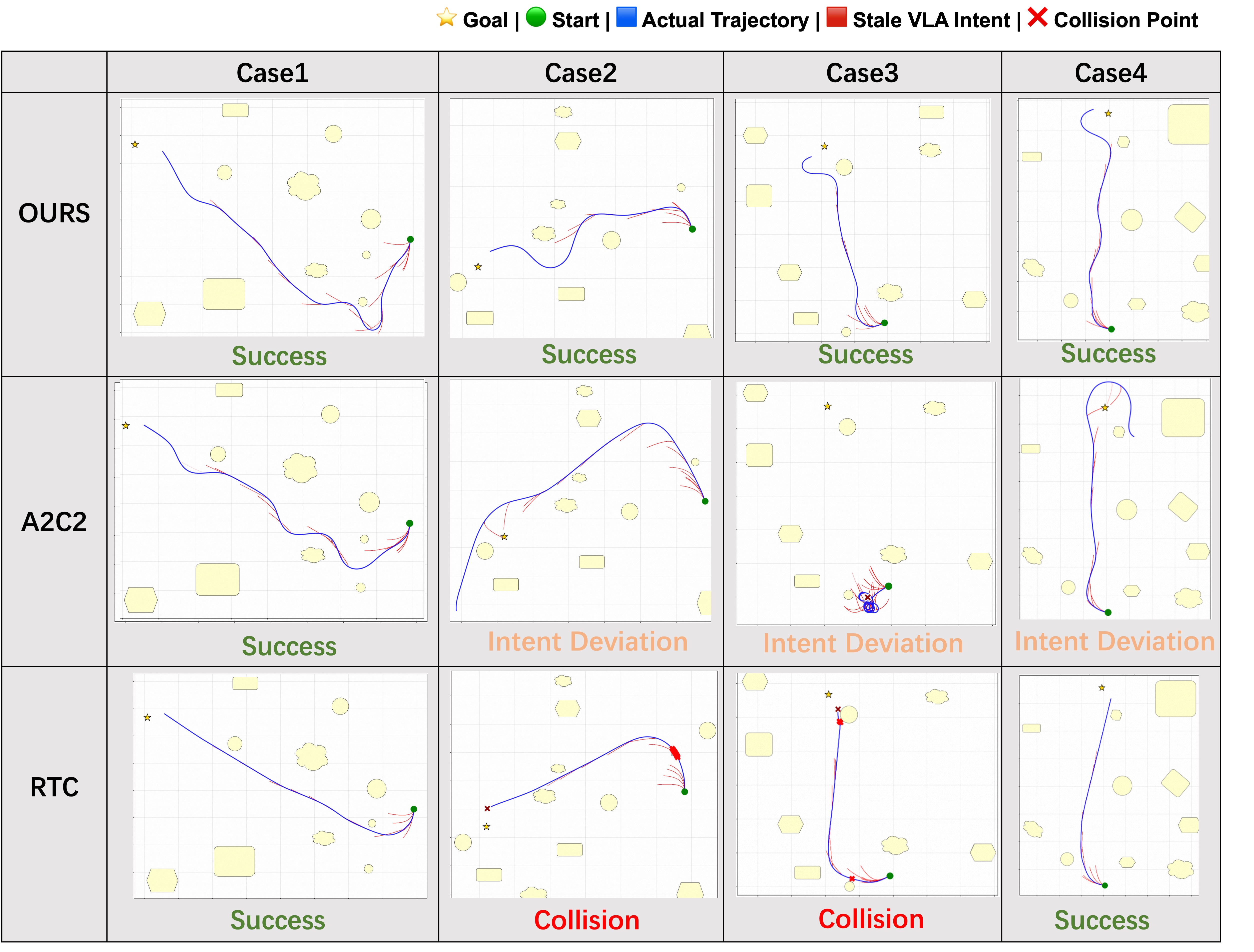} 
    \caption{\textbf{Qualitative comparison of executed trajectories under the Mixed Degradation network condition.} We visualize the actual executed robot trajectories (blue solid lines) and the realigned/stale VLA intents (red thin lines) across four highly challenging dynamic scenarios. \textbf{RTC} generates extremely smooth curves but blindly guides the robot to crash into dynamic obstacles. \textbf{A2C2} exhibits severe \textit{Intent Deviation}, causing the robot to oscillate or completely distort the intended path. \textbf{AsyncShield (Ours)} achieves a perfect trade-off: it rigorously restores the VLA's original intent in free space, and autonomously deviates to ensure safety when the intent becomes dangerous.}
    \label{fig:traj_comparison}
\end{figure*}

\noindent\textbf{Environment and Task Setup:}
We construct a 3D navigation environment in OmniSafe. The robot navigates to a target point based on VLA-generated local waypoints. To evaluate the system's obstacle avoidance and error correction capabilities, we randomize the test scenarios:
\begin{itemize}
    \item \textbf{Spatial Complexity:} Each episode is conducted in a randomly generated $10\text{m} \times 10\text{m}$ map.
    \item \textbf{Mixed Obstacle Interference:} We deploy 7 static and 4 dynamic wandering obstacles to evaluate physical safety responses.
    \item \textbf{Cross-Embodiment Dynamics Randomization:} We apply a $\pm 40\%$ domain randomization to the robot's kinematic constraints (velocity and acceleration limits) to verify the RL Adapter's universality.
\end{itemize}

\noindent\textbf{Network Conditions \& Degradation Modeling:}
Assuming the cloud VLA generates waypoints at roughly $3\text{ Hz}$, we evaluate the edge execution under two communication profiles:
\begin{itemize}
    \item \textit{Ideal (Fast \& Stable Update):} A reliable network with a deterministic ${\sim}200\text{ ms}$ delay and zero packet loss.
    \item \textit{Non-ideal (Mixed Degradation):} A mixed degradation model emulating unreliable wireless communication. It applies three independent perturbations:
    \begin{itemize}
        \item \textbf{Heavy-Tail Latency:} To emulate irregular cloud-side queuing and network delays, the one-way cloud-to-edge delay $\Delta t$ is sampled from a mixture distribution: 
        $\Delta t \sim (1-q) \cdot \mathcal{U}(0.15, 0.25) + q \cdot \mathcal{U}(0.5, 1.5)\text{ s}$, 
        where $q = 0.1$ is the probability of a latency spike.
        \item \textbf{Stochastic Packet Loss:} To simulate signal interference, each transmitted message is subject to an independent Bernoulli drop with probability $p_{\text{loss}} = 0.15$.
        \item \textbf{Transient Outages (Black-outs):} To emulate physical signal dead zones, we define an outage time ratio $r_{\text{out}} = 0.05$, where during randomly sampled continuous segments $D \sim \mathcal{U}(1.0, 2.0)\text{ s}$, the packet loss rate is temporarily forced to $100\%$.
    \end{itemize}
\end{itemize}

\noindent\textbf{Baselines:}
We adapt two state-of-the-art asynchronous execution strategies to the embodied navigation domain for comparison:
\begin{itemize}
    \item \textbf{Naive (Direct Execution):} Directly executes the VLA-generated local waypoints as current commands without any spatio-temporal alignment.
    \item \textbf{RTC~\cite{black2025realtimeexecutionactionchunking} (Real-Time Chunking):} Focuses on the smooth temporal transition of action chunks. We adapt its core mechanism to smoothly stitch consecutive VLA waypoint chunks to eliminate trajectory jitter.
    \item \textbf{A2C2~\cite{sendai2025leaveobservationbehindrealtime} (Asynchronous Action Chunk Correction):} Utilizes a lightweight, high-frequency RL correction head to output residual actions based on the latest observation.
\end{itemize}

\vspace{-2mm} 
\noindent\textbf{Evaluation Metrics:}
\vspace{-1mm} 
\begin{itemize}
    \item \textbf{SR (Success Rate):} Overall task completion rate.
    \item \textbf{CTE (Cross-Track Error):} Episode-weighted trajectory tracking error as defined in \cite{penumarti2024globaluncertaintyawareplanningmagnetic}, measuring the fidelity to the VLA's original intent.
    \item \textbf{RER (Risk Exposure Rate):} The percentage of time steps where the robot is exposed to high-collision-risk regions ($d_{\min} < d_{\text{risk}}$), reflecting physical safety.
    \item \textbf{TTG (Time-to-Goal):} Navigation efficiency as utilized in \cite{rajagopal2025drnavsemanticgeometricrepresentations}. To eliminate survivorship bias, TTG is calculated strictly on the shared-success subset across all methods.
\end{itemize}


\subsection{Robustness and Safety Analysis}

\begin{table*}[t]
\centering
\caption{Ablation study under the same evaluation protocol as Table~\ref{tab:main_results}. \textit{Note:} `w/o Safety Constraints' disables the Lagrangian safety optimization during training/execution.}
\label{tab:ablation}
\begin{tabular}{l ccc ccc}
\toprule
\multirow{2}{*}{\textbf{Method Variant}} & \multicolumn{3}{c}{\textbf{Ideal (Fast Update)}} & \multicolumn{3}{c}{\textbf{Non-ideal (Mixed Degradation)}} \\
\cmidrule(lr){2-4} \cmidrule(lr){5-7}
& SR (\%) $\uparrow$ & CTE (m) $\downarrow$ & RER (\%) $\downarrow$ & SR (\%) $\uparrow$ & CTE (m) $\downarrow$ & RER (\%) $\downarrow$ \\
\midrule
\textbf{AsyncShield (Full)} & \textbf{80.0} & 0.717 & \textbf{1.2} & \textbf{76.7} & 0.725 & \textbf{1.3} \\
\midrule
w/o Temporal Alignment   & 53.3 & 0.915 & 1.6 & 36.7 & 1.194 & 3.6 \\
w/o RL Adapter & 66.7 & 1.232 & 1.3 & 53.3 & 1.443 & 1.4 \\
w/o Safety Constraints   & 40.0 & 0.681 & 4.2 & 23.3 & 0.692 & 4.7 \\
\bottomrule
\end{tabular}
\end{table*}

Statistical results across 600 evaluation episodes (summarized in Table~\ref{tab:main_results}) reveal several critical insights regarding asynchronous navigation:

\subsubsection{Overall Task Completion and Safety}
Across both network conditions, AsyncShield achieves the highest task completion (SR: $80.0\% \rightarrow 76.7\%$) and maintains consistently low risk exposure (RER: $1.2\% \rightarrow 1.3\%$). Baseline methods degrade substantially under injected stochastic latency (e.g., A2C2 drops from $56.7\%$ to $43.3\%$, and RTC from $40.0\%$ to $30.0\%$). This discrepancy stems from domain differences: RTC and A2C2 excel in manipulation tasks where intent smoothing and local residual fitting suffice. However, in mobile navigation, large-scale spatial misalignment leads to critical collisions. AsyncShield absorbs this misalignment via $SE(2)$ geometric transformations and an RL adapter, providing robustness to stochastic latency.

\subsubsection{The CTE Paradox: Tracking Fidelity vs. Safety}
A counter-intuitive phenomenon emerges under the Ideal condition: RTC achieves the lowest trajectory tracking error (CTE = $0.673\text{~m}$), yet exhibits a low SR ($40.0\%$) and high RER ($3.0\%$). This reveals a critical navigation paradigm: \textit{blind smooth tracking does not equate to safety}. While RTC seamlessly stitches action chunks, it smoothly executes stale VLA intents, directing the robot toward dynamic obstacles. AsyncShield records a slightly higher CTE ($0.717\text{~m}$) because the PPO-Lagrangian mechanism proactively deviates from unsafe intents for obstacle avoidance. This demonstrates that our ``intent-safety decoupling'' effectively trades minor tracking fidelity for substantial survival rate gains.

\subsubsection{Efficiency on Shared-Success Episodes}
Under the Ideal condition, the Naive method appears fastest (TTG = $7.92\text{~s}$). However, this manifests \textit{survivorship bias}---with an SR of only $20.0\%$, Naive strictly succeeds in trivial, obstacle-free scenarios. Evaluated strictly on the shared-success subset of complex tasks, latency inflates TTG across all methods. AsyncShield exhibits the smallest efficiency degradation ($8.87\text{~s} \rightarrow 9.29\text{~s}$), outperforming A2C2 ($9.64\text{~s}$), RTC ($9.81\text{~s}$), and Naive ($10.25\text{~s}$), demonstrating robust decision-making under adverse latency conditions.

\subsubsection{Qualitative Analysis}
To intuitively understand behavioral differences under the severe Mixed Degradation network, we visualize the executed trajectories (blue solid lines) and realigned/stale VLA intents (red thin lines) across four challenging dynamic scenarios in Fig.~\ref{fig:traj_comparison}. The visualizations highlight the failure modes of the baselines:
\begin{itemize}
    \item \textbf{The Pitfall of Blind Smoothness (RTC):} In Case 2 and Case 3, RTC successfully eliminates trajectory oscillation, generating a smooth curve. However, lacking explicit spatial realignment and obstacle perception, RTC smoothly but blindly steers the robot into dynamic obstacles. This matches its high RER.
    \item \textbf{The Collapse of Residual Fitting (A2C2):} In Case 3 and Case 4, A2C2 exhibits severe Intent Deviation. Without an explicit geometric anchor, a pure RL residual network struggles to implicitly map large time delays into spatial coordinate corrections. This cumulative error causes the robot to exhibit severe oscillation or completely distort the intended path.
    \item \textbf{Adaptive Trade-off between Geometry and Safety (AsyncShield):} Our method demonstrates exceptional robustness. Driven by the PPO-Lagrangian optimization, the system dynamically adjusts the weights between intent-tracking rewards and safety costs. Consequently, AsyncShield rigorously restores the VLA's original intent in obstacle-free spaces, and autonomously prioritizes collision avoidance when stale intents become dangerous, seamlessly blending task fidelity with physical safety.
\end{itemize}

\subsection{Ablation Studies}



To validate the necessity of AsyncShield's core architectural designs, we conduct targeted ablation experiments (see Table~\ref{tab:ablation}):

\begin{itemize}
    \item \textbf{w/o Temporal Alignment (Naive Grafting):} Removing the temporal pose buffer forces the agent to rigidly graft stale local waypoints onto the current ego frame. Under Delayed latency, SR plummets from $76.7\%$ to $36.7\%$, and CTE deteriorates to $1.194\text{~m}$. This strongly corroborates our core premise, ``Latency is Geometry'': under severe stochastic latency, relying on backend RL networks to implicitly fit large-scale coordinate frame discrepancies is highly inefficient without explicit spatial realignment.
    


    \item \textbf{w/o RL Adapter:} Replacing the RL policy with a classical dynamic window approach (DWA) planner. While DWA effectively avoids obstacles ($66.67\%$ SR under Ideal), its success drops to $53.30\%$ under latency, accompanied by severe trajectory deviation (CTE $> 1.2$~m). This reveals that traditional cost-based planners struggle to balance stale intents with safety, surviving only by aggressively abandoning the VLA's guidance. Conversely, our CMDP-based RL adapter achieves a tuning-free, optimal trade-off between intent fidelity and physical safety.
    
    \item \textbf{w/o Safety Constraints:} Removing the Lagrangian-based obstacle penalty relies purely on flow-tracking rewards. Consequently, the CTE decreases as the agent rigidly adheres to the VLA intent, but RER skyrockets, resulting in task failures due to direct collisions. This inverse validation underscores the indispensable role of the PPO-Lag mechanism as a hard safety baseline in asynchronous heterogeneous navigation.
\end{itemize}

\subsection{Cross-Embodiment Validation in Simulation}
To validate the zero-shot cross-embodiment capability of AsyncShield, we directly deploy the base policy onto two morphologically distinct agents in the OmniSafe simulator: \textbf{Doggo} (a quadruped robot) and \textbf{Racecar} (a vehicle with Ackermann steering constraints), designing significantly different kinematic parameters for each category of robot. 

Specifically, to eliminate the discrepancy in physical volumes across different embodiments without the need for retraining, we introduce a \textbf{Collision Radius Inflation} mechanism. By subtracting a specific inflation compensation value directly from the raw LiDAR scan distances at the perception level (i.e., equivalently pushing obstacles closer in the observation space), the base policy can zero-shot adapt to physical entities with larger collision volumes.

As shown in Table \ref{tab:cross_embodiment}, the differences in Success Rate (SR) and Risk Exposure Rate (RER) across all variants are minimal. This proves that our method can safely and effectively generalize to other embodiments.

\begin{table}[h]
\centering
\caption{Zero-Shot Cross-Embodiment Performance}
\label{tab:cross_embodiment}
\begin{tabular}{lcc}
\toprule
\textbf{Embodiment Variant} & \textbf{SR (\%)} $\uparrow$ & \textbf{RER (\%)} $\downarrow$ \\
\midrule
\textbf{Doggo A}  & 78.00 & 1.25 \\
\textbf{Doggo B}  & 76.00 & 1.31 \\
\midrule
\textbf{Racecar A} & 79.00 & 1.20 \\
\textbf{Racecar B} & 76.00 & 1.29 \\
\bottomrule
\end{tabular}
\end{table}

\subsection{Real-world Hardware Deployment and Zero-Shot Transfer}

To validate the ``plug-and-play'' feasibility and zero-shot transferability, we deployed AsyncShield on a Unitree Go2 quadruped robot. The system adopts an edge-cloud architecture: the onboard computer processes 2D LiDAR and odometry at high frequencies, while a cloud GPU runs the VLA models, communicating via real-world Wi-Fi with a baseline round-trip latency of approximately 200ms.

\noindent\textbf{Experimental Setup and Multi-Task Validation:} 
We selected three SOTA VLA models covering distinct tasks to verify the generalizability of AsyncShield: \textbf{SocialNav}~\cite{chen2025socialnavtraininghumaninspiredfoundation} for point-goal navigation, \textbf{TrackVLA}~\cite{wang2025trackvla} for person-following, and \textbf{Nav-$R^2$}~\cite{xiang2025navr2dualrelationreasoninggeneralizable} for object-goal navigation. For each model, we evaluated performance across four challenging scenarios: \textit{Dynamic Crowds}, \textit{Narrow Corridors}, \textit{Doorway Entry/Exit}, and \textit{Extreme Network Jitter}. In the jitter scenario, we artificially injected stochastic latencies ranging from 500ms to 1500ms on top of the baseline delay. We conducted 5 trials per scenario, totaling 20 cases for each VLA model.

\begin{table}[h]
\centering
\small
\caption{Real-world Success Rates (20 trials each) demonstrating plug-and-play compatibility across different cloud VLA models.}
\label{tab:real_world_vla}
\begin{tabular}{lcc}
\toprule
\textbf{Cloud VLA Model} & \textbf{Direct VLA} & \textbf{VLA + AsyncShield} \\
\midrule
SocialNav (Point-goal) & 6/20 (30\%) & \textbf{17/20 (85\%)} \\
TrackVLA (Following)  & 8/20 (40\%) & \textbf{16/20 (80\%)} \\
Nav-$R^2$ (Object-goal) & 5/20 (25\%) & \textbf{18/20 (90\%)} \\
\bottomrule
\end{tabular}
\end{table}

\noindent\textbf{Results and Qualitative Analysis:} 
As shown in Table \ref{tab:real_world_vla}, due to the severe cloud-to-edge latency, naive asynchronous execution (\textit{Direct VLA}) suffers catastrophic performance degradation (25\%--40\% SR) in challenging scenarios, frequently leading to collisions caused by stale commands. Integrating our module immediately restores robust performance (80\%--90\% SR) without any VLA fine-tuning.

In dynamic environments and narrow spaces, AsyncShield allows the robot to autonomously deviate from stale global intents based on real-time local perception. For instance, during doorway traversal, it recalculates corrective trajectories in real-time to prevent lateral drift or wall collisions induced by command lag. Furthermore, in scenarios with induced high network latency, the robot exhibits no oscillations or freezing; instead, it safely proceeds along the primary intent direction while maintaining active obstacle avoidance, demonstrating remarkable resilience to communication failures.

Synthesizing the aforementioned simulation and real-world experiments, the cross-embodiment generalization capability validated in simulation mutually corroborates the hardware deployment performance on the real-world quadruped robot. This demonstrates that AsyncShield can serve as a universal plug-and-play edge module, safely and robustly enabling the real-world physical deployment of cloud-based VLA models without requiring any fine-tuning.

\section{CONCLUSION}
\label{sec:conclusion}
This paper proposes AsyncShield, a plug-and-play asynchronous control framework designed for cloud-based Vision-Language-Action (VLA) models in mobile navigation tasks. To address the spatio-temporal misalignment caused by cloud-to-edge latency, we discard traditional black-box time-series prediction in favor of a deterministic physical white-box spatial mapping. We restore the original intent of the VLA through geometric transformations and formulate the edge adaptation as a CMDP. Utilizing the PPO-Lagrangian algorithm, we develop a reinforcement learning adapter to achieve an adaptive dynamic trade-off between tracking the true VLA intent and responding to high-frequency LiDAR obstacle avoidance hard constraints.

Furthermore, benefiting from a standardized universal sub-goal interface, domain randomization, and perception-level adaptation (i.e., Collision Radius Inflation), AsyncShield demonstrates zero-shot cross-embodiment generalization capabilities. Simulation and real-world experiments show that, without fine-tuning any cloud-based foundation model weights, the framework exhibits stable generalization capability and effectively improves the success rate and physical safety of asynchronous navigation.

Future work will explore extending this analytical geometric re-projection mechanism to more complex 3D unstructured environments, and investigate the introduction of lightweight multimodal local perception models at the edge to further enhance the system's safety redundancy and robustness in dynamic scenarios.

\bibliographystyle{IEEEtran}
\bibliography{IEEEabrv, refs}

\end{document}